\documentclass[letterpaper,10pt,conference]{ieeeconf}  %

\IEEEoverridecommandlockouts                              

\usepackage{epsfig} 
\usepackage{times} 
\usepackage{amsmath} 
\usepackage{amssymb}  
\usepackage{amsfonts}
\usepackage{booktabs}
\usepackage{algorithm}
\usepackage{algpseudocode}
\usepackage{bm}  
\usepackage{url}
\usepackage{hyperref}
\usepackage[dvipsnames]{xcolor}
\usepackage{multirow}
\usepackage{comment}
\usepackage{subcaption}
\usepackage{cite}

\definecolor{commentGreen}{rgb}{0,0.5,0.05}

\newcommand{\best}[1]{\textcolor{Green}{\bm{#1}}}
\newcommand{\second}[1]{\textcolor{blue}{\underline{#1}}}

\title{Curriculum Multi-Task Self-Supervision Improves Lightweight Architectures for Onboard Satellite Hyperspectral Image Segmentation\vspace{-0.3cm}}

\author{%
Hugo Carlesso$^{1,2}$, Josiane Mothe$^{1,3}$, Radu Tudor Ionescu$^{4}$\\
\hspace{-0.1em}$^{1}$Univ. Toulouse, IRIT, France\hspace{0.4em} $^{2}$CNRS, France\hspace{0.4em} $^{3}$CLLE, CNRS, France\hspace{0.4em} $^{4}$University of Bucharest, Romania\\
{\tt\small hugo.carlesso@cnrs.fr, josiane.mothe@irit.fr, raducu.ionescu@gmail.com}\vspace{-0.5cm}
\thanks{*This project has received financial support from the CNRS through the MITI interdisciplinary programs. This work was also supported by a grant of the Ministry of Research, Innovation and Digitization, CNCS -
UEFISCDI, project number PN-IV-P1-PCE-2023-0354, within PNCDI IV.}
}

\begin{document}

\maketitle
\thispagestyle{empty}
\pagestyle{empty}

\begin{abstract}
Hyperspectral imaging (HSI) captures detailed spectral signatures across hundreds of contiguous bands per pixel, being indispensable for remote sensing applications such as land-cover classification, change detection, and environmental monitoring. Due to the high dimensionality of HSI data and the slow rate of data transfer in satellite-based systems, compact and efficient models are required to support onboard processing and minimize the transmission of redundant or low-value data. To this end, we introduce a novel curriculum multi-task self-supervised learning (CMTSSL) framework designed for lightweight architectures for HSI analysis. CMTSSL integrates masked image modeling with decoupled spatial and spectral jigsaw puzzle solving, guided by a curriculum learning strategy that progressively increases data difficulty during self-supervision. This enables the encoder to jointly capture fine-grained spectral continuity, spatial structure, and global semantic features. Unlike prior dual-task SSL methods, CMTSSL simultaneously addresses spatial and spectral reasoning within a unified and computationally efficient design, being particularly suitable for training lightweight models for onboard satellite deployment. We validate our approach on four public benchmark datasets, demonstrating consistent gains in downstream segmentation tasks, using architectures that are over $\mathbf{16,000\times}$ lighter than some state-of-the-art models. These results highlight the potential of CMTSSL in generalizable representation learning with lightweight architectures for real-world HSI applications. Our code is publicly available at: \url{https://github.com/hugocarlesso/CMTSSL}.
\end{abstract}

\setlength{\abovedisplayskip}{2.0pt}
\setlength{\belowdisplayskip}{2.0pt}

\section{Introduction}
\vspace{-0.1cm}
Hyperspectral imaging (HSI) provides detailed spectral signatures across hundreds of contiguous bands per pixel, creating informative \emph{data cubes} that enable fine-grained material recognition and environmental monitoring. This makes HSI an invaluable tool in satellite-based Earth observation, where robust scene understanding is essential for tasks such as land-cover classification, change detection, and resource monitoring~\cite{li_deep_2019, shahmoradi_comprehensive_2020, wendel2016self, yang2021enhanced, zhang2025hybrid}. The operational constraints of space-borne platforms pose unique challenges for HSI data processing. Satellites are required to perform onboard inference under strict computational and energy budgets, often relying on edge devices with limited memory and processing capabilities. Additionally, downlink bandwidth is limited \cite{justo_semantic_2025}, making it crucial to transmit only relevant or high-quality data, e.g.~excluding scenes with excessive cloud cover. These constraints demand lightweight and efficient models, capable of operating on edge.

\begin{figure}[!htbp]
  \centering
  \includegraphics[width=1.0\linewidth]{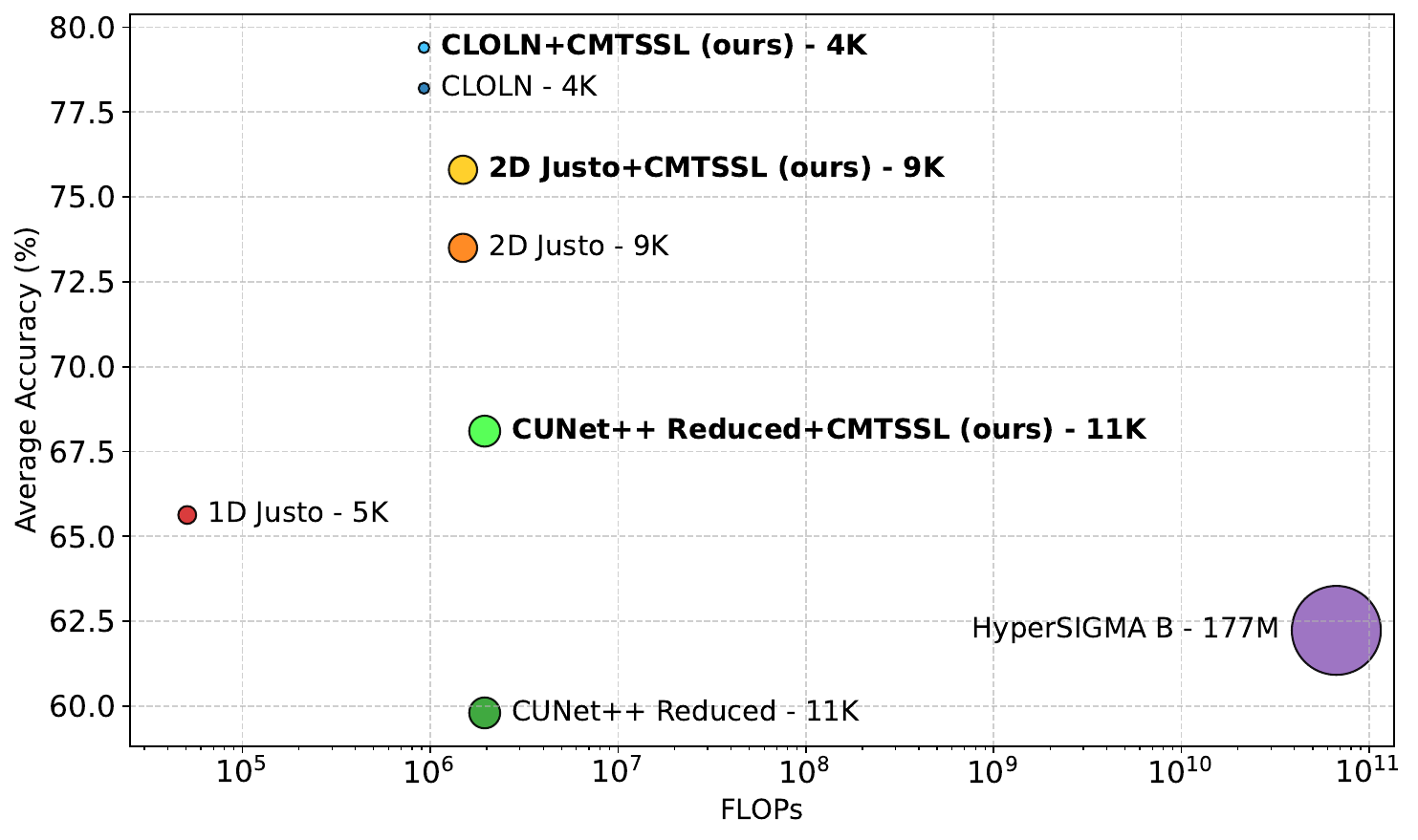}
  \vspace{-0.7cm}
  \caption{Average accuracy vs.~number of floating point operations (FLOPs) for various HSI architectures, including both lightweight models (1D Justo \cite{justo_semantic_2025}, 2D Justo \cite{justo_semantic_2025}, CLOLN \cite{li2024channel}, CUNet++ Reduced \cite{justo_semantic_2025}) and foundation models (HyperSIGMA-B \cite{wang_hypersigma_2025}). Circle areas indicate parameter counts. Performance is reported on the Pavia University dataset \cite{noauthor_hyperspectral_nodate}. Our CMTSSL boosts performance of lightweight models, without affecting model size and FLOPs.}
  \vspace{-1.9em}
   \label{fig:teaser}
\end{figure}

An orthogonal limitation is that acquiring high-quality labels for supervised training on hyperspectral images remains prohibitively expensive. 
Due to the scarcity of annotated large-scale HSI datasets, researchers in the community embraced self-supervised learning (SSL) as a means to harness the rich spatial-spectral structure of unlabeled data~\cite{wang_self-supervised_2022}. SSL frameworks employ label-free pretext tasks to learn meaningful representations, which can further be fine-tuned on downstream tasks with limited supervision~\cite{ericsson_how_2021}. For HSI, such approaches have demonstrated superior performance over traditional supervised baselines in classification, segmentation, and anomaly detection~\cite{wang_hypersigma_2025}. Among SSL frameworks, contrastive learning and masked image modeling (MIM)~\cite{hondru2025masked} have been widely adopted. Yet, these approaches suffer from key limitations. Contrastive methods~\cite{zhu_sc-eadnet_2022, huang_3-d-swin_2022} may fail to capture fine-grained details, MIM pipelines~\cite{scheibenreif_masked_2023} can struggle with high-level semantic separability, and discriminative frameworks \cite{shu2023learning,yang_self-supervised_2022} often offer limited generalization. Furthermore, SSL methods are not optimized for lightweight architectures designed for onboard satellite processing.

To overcome these challenges, we introduce CMTSSL, a \textbf{c}urriculum \textbf{m}ulti-\textbf{t}ask \textbf{s}elf-\textbf{s}upervised \textbf{l}earning framework for lightweight architectures explicitly designed for onboard hyperspectral image segmentation. CMTSSL integrates MIM with decoupled spatial and spectral jigsaw puzzle solving (JPS) tasks \cite{Wang_video_2022}, within a curriculum learning paradigm \cite{bengio2009curriculum}. We find that 3D gradient magnitudes of HSI data cubes are highly correlated with the capability of solving self-supervised tasks on the respective cubes. Consequently, we design a curriculum learning procedure, organizing the samples into progressively difficult batches, from easy (low gradient magnitude) to hard (high gradient magnitude). By integrating multiple tasks and progressively increasing the difficulty of samples during self-supervision, our framework encourages the joint encoder to gradually learn robust complementary representations spanning spectral continuity, spatial arrangement, and global semantic structure. Our novel learning framework boosts the performance of lightweight architectures without increasing model size or the number of floating point operations (FLOPs), as shown in Figure \ref{fig:teaser}. While the common approach is to train lightweight models via supervised learning \cite{justo_semantic_2025,li2024channel}, we introduce CMTSSL as an encoder-agnostic pretraining stage, to be carried out before the conventional supervised training stage. 

We carry out extensive experiments on four standard HSI datasets: Pavia University~\cite{noauthor_hyperspectral_nodate}, Pavia Center~\cite{noauthor_hyperspectral_nodate}, WHU-HI Hanchuan~\cite{hu2020whu}, and HYPSO~\cite{justo_open_2023}. The reported results confirm that introducing the CMTSSL stage is beneficial. Moreover, we obtain a new state-of-the-art average accuracy ($93.5\%$) on the large-scale HYPSO benchmark. Hence, our contribution lays the groundwork for robust edge-deployable solutions for satellite-based HSI analysis.

In summary, our contribution is threefold:
\begin{itemize}
    \item We propose a novel curriculum multi-task learning framework for HSI data, which encourages lightweight encoders to acquire robust representations by learning to solve decoupled spatial-spectral jigsaw puzzles and reconstruct masked data cubes.     
    \item We adapt the jigsaw puzzle solving framework to hyperspectral images, decoupling the task across spatial and spectral dimensions, while using a joint encoder.
    \item We introduce a novel curriculum learning strategy for HSI data based on 3D gradient magnitudes, which progressively increases sample difficulty to guide the multi-task self-supervised learning.
\end{itemize}

\section{Related Work}
\label{Sec:related work}

\subsection{Self-Supervised Learning} 

Self-supervised models learn data representations without human-annotated labels. Instead, pretext tasks serve to generate supervisory signals directly from the data, such as learning the arrow of time \cite{Wei-CVPR-2018}, solving jigsaw puzzles \cite{noroozi-eccv-2016,Chen-CVPR-2021}, reconstructing masked information \cite{hondru2025masked,he-cvpr-2022,xie-cvpr-2022}, or contrasting augmented views \cite{Chen-ICML-2020, Grill-NeurIPS-2020, He-CVPR-2020}. SSL enables effective training with limited labeled data and supports domain-specific pretraining, thereby mitigating domain gaps~\cite{song_self-supervised_2022}. SSL is particularly promising in remote sensing \cite{ayuba2025specbpp}, where large volumes of unlabeled Earth observation data are available, yet the cost to manually annotate this data is high. Recent advances demonstrate strong potential for both multispectral and hyperspectral imagery. For instance, TerraMind \cite{jakubik2025terramind} is a large-scale multimodal foundation model for Earth observation, while HyperSIGMA \cite{wang_hypersigma_2025} and SpectralEarth \cite{braham2025spectralearth} leverage large HSI datasets (450K and 539K images, respectively) to train foundation models with high transferability. SSL methods in hyperspectral imaging often employ contrastive learning with spectral augmentations \cite{zhu_sc-eadnet_2022, huang_3-d-swin_2022}, MIM for spatial-spectral reconstruction \cite{scheibenreif_masked_2023}, and auxiliary discriminative tasks enforcing spatial-spectral reasoning \cite{yang_self-supervised_2022}. 

SSL exhibits strong potential in remote sensing, but its effectiveness heavily depends on the choice of a single pretext task. 
A given objective may capture some aspects of the data, such as invariance to augmentation or local spatial structures, while neglecting others, leading to suboptimal representations. To overcome this limitation, some recent frameworks \cite{muhtar2023cmid,zhang_unified_2025, chen_spectral-spatial_2025} integrated multiple SSL tasks to obtain more robust representations. For example, CMID \cite{muhtar2023cmid} fuses contrastive learning with masked image modeling through self-distillation, yielding representations that are both semantically rich and spatially grounded. However, solving multiple tasks can lead to an arduous learning process, requiring a careful balancing between learning objectives \cite{liu2021conflict}. Unlike existing multi-task SSL frameworks for HSI, we introduce a curriculum learning strategy to gradually increase task complexity by adding more difficult images, guiding the learning process towards the desired objective.

\vspace{-0.12cm}
\subsection{Curriculum Learning}
\vspace{-0.1cm}
Curriculum learning~\cite{bengio2009curriculum} is a training strategy inspired by human learning: models first learn from simple examples before tackling more difficult ones. Replacing random data ordering with an easy-to-hard progression can speed up convergence and enhance generalization. A typical curriculum learning framework consists of two main elements: a difficulty criterion, which ranks training samples from easy to hard, and a scheduler, which determines when increasingly difficult examples are introduced during training. Curriculum learning can be applied at various levels, including the data \cite{bengio2009curriculum, chen2015webly, soviany2021curriculum}, the task \cite{Caubriere-INTERSPEECH-2019}, the model capacity \cite{karras2017progressive,croitoru2025learning,Sinha-NIPS-2020}, and even the optimization objective \cite{soviany2022curriculum}.

Curriculum learning has also been applied in SSL frameworks based on masked information modeling \cite{madan-wacv-2024,seth2024eh}. Fo instance, CL-MAE \cite{madan-wacv-2024} integrates a learnable module that selects image patches for masking based on the difficulty to reconstruct the masked patches. However, the module and the masked autoencoder need to be trained in alternating steps. EH-MAM \cite{seth2024eh} is an audio SSL framework that employs a teacher model to decide which acoustic frames to mask. Unlike such frameworks \cite{madan-wacv-2024,seth2024eh}, our strategy exploits the correlation between SSL tasks and image gradient magnitudes to create a data-level curriculum, requiring no extra model to generate the curriculum. To the best of our knowledge, a data-level curriculum based on 3D gradients for SSL on HSI data has not been explored in previous literature. Our difficulty criterion stems from our observation that gradient magnitudes are well correlated to the SSL tasks integrated into the proposed framework.

\vspace{-0.1cm}
\subsection{Lightweight HSI Architectures}
\vspace{-0.1cm}

\begin{figure*}[!ht]
  \centering
  \includegraphics[width=0.97\textwidth]{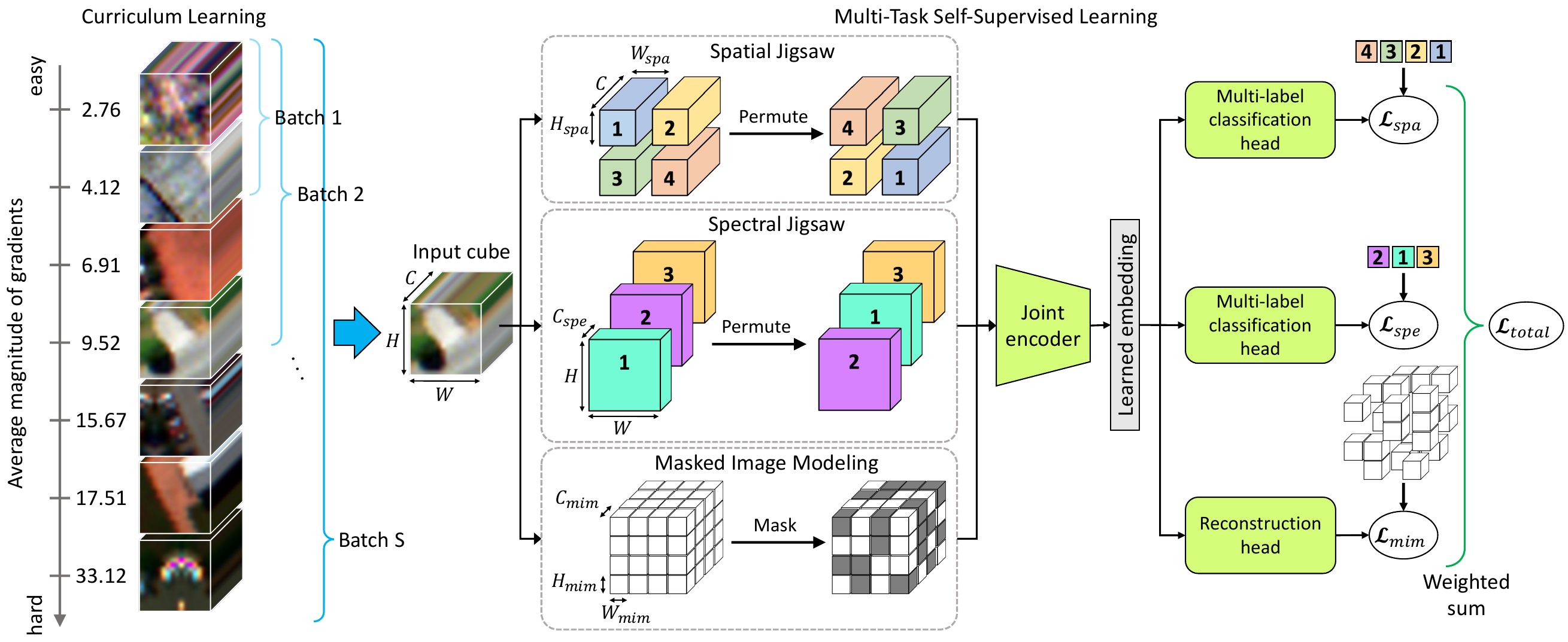}
  \vspace{-0.2cm}
  \caption{Curriculum multi-task self-supervised learning (CMTSSL) pipeline. Input HSI cubes are sorted according to their 3D gradient magnitudes and divided into $S$ curriculum batches. To generate the pretext tasks, each input cube undergoes three parallel transformations: random spatial permutation, spectral permutation, and patch masking. The transformed cubes are encoded via a shared encoder and routed to task-specific heads, namely spatial permutation prediction, spectral permutation prediction, and masked image modeling (MIM). The model is jointly optimized via a weighted loss. Best viewed in color.}
  \label{fig:full_pipeline}
  \vspace{-1.5em}
\end{figure*}

In remote sensing, especially for onboard processing, lightweight architectures that preserve high accuracy are of growing interest \cite{justo_semantic_2025,li2024channel,justo2025hyperspectral}. Li et al.~\cite{li2024channel} introduced the Channel-Layer-Oriented Lightweight Spectral-Spatial Network (CLOLN), which explicitly models both spectral and spatial interactions, while minimizing parameter and computational overhead.
Justo et al.~\cite{justo_semantic_2025} proposed an entire lineup of lightweight 1D or 2D CNNs for onboard HSI segmentation. For example, 1D Justo-LiuNet~\cite{justo2025hyperspectral} is a compact 1D CNN focused on spectral feature extraction, which was deployed on a satellite. Despite having only about $4,500$ parameters, it outperforms more complex models, such as fast vision transformers (FastViT) \cite{anasosalu_vasu_fastvit_2023}. 
Existing lightweight models suitable for deployment in scenarios with severe resource constraints, such as in-orbit processing, are all based on supervised learning. To the best of our knowledge, we are the first to show that multi-task SSL is beneficial for such lightweight models.

\section{Method}
\label{Method}

We propose a multi-task self-supervised learning (MTSSL) framework that integrates decoupled spatial and spectral jigsaw puzzle solving tasks with masked image modeling (see Figure \ref{fig:full_pipeline}). %
We hypothesize that integrating multiple SSL tasks can enable the shared encoder to learn complementary cues about local continuity, structural arrangement, and global semantics within a single paradigm, essentially leading to a more robust representation (embedding). However, the integration of multiple objectives can make learning difficult and can even result in worse performance for each task than in single-task SSL \cite{liu2021conflict}. To mitigate this issue, we propose a curriculum learning schedule constructed around the \emph{saliency} of 3D input images, which is estimated via 3D image gradient magnitudes. This approach is based on our preliminary findings suggesting that the difficulty of reconstructing masked patches or rearranging visible patches increases along with the saliency level of the input samples. Unlike heuristic loss balancing or auxiliary difficulty estimators, our scheme is data-driven, architecture-agnostic, and requires no extra model, mitigating negative transfer, while preserving the benefits of multiple SSL objectives. We formally present our curriculum multi-task self-supervised learning (CMTSSL) framework in Algorithm \ref{alg:cmtssl}. We continue with a detailed description of each component, while referring to the corresponding steps of the algorithm along the way.

\vspace{-0.1cm}
\subsection{Mathematical Preliminaries and Notations}
\label{notation}
\vspace{-0.1cm}
In the remainder of this work, we use bold letters to denote vectors, matrices and tensors, round letters to denote sets, lowercase letters to denote neural blocks, and uppercase letters to denote scalar values. We use $\odot$ to represent the convolution operation, and $\mathbf{X}^{*P}$ to represent the element-wise raise of matrix $\mathbf{X}$ to power $P$. We denote a hyperspectral image (data cube) as $\mathbf{I} \in \mathbb{R}^{H \times W \times C}$, where $H$ is the height, $W$ is the width and $C$ is the number of channels (bands).

\newlength{\textfloatsepsave} 
\setlength{\textfloatsepsave}{\textfloatsep}
\setlength{\textfloatsep}{4pt}

\begin{algorithm}[!t]
    \caption{CMTSSL Framework}
    \label{alg:cmtssl}
    \small{
\begin{algorithmic}[1]
\Require 
    $\mathcal{D} = \{\mathbf{I}_i\}_{i=1}^N$ - set of $N$ training images; 
    $\mathbf{K}_x, \mathbf{K}_y, \mathbf{K}_y$ - filters to compute gradients;
    $S$ - number of curriculum batches; 
    $K$ - initial number of epochs; 
    $F$ - growth factor for the number of epochs; 
    $E$ - number of epochs for current batch;
    $f_{\boldsymbol{\theta}}$ - joint encoder with weights $\boldsymbol{\theta}$; 
    $h_{\boldsymbol{\phi}}^{\text{spa}}$ - spatial jigsaw multi-label classification head with weights $\boldsymbol{\phi}$;
    $h_{\boldsymbol{\beta}}^{\text{spe}}$ - spectral jigsaw multi-label classification head with weights $\boldsymbol{\beta}$;
    $h_{\boldsymbol{\gamma}}^{\text{mim}}$ - masked patch reconstruction head with weighs $\boldsymbol{\gamma}$; 
    $M$ - number of patches to be masked; 
    $\alpha_{\text{spa}}, \alpha_{\text{spe}}, \alpha_{\text{mim}}$ - weights used for the individual losses; $\eta$ - learning rate.
\Ensure
$f^{\boldsymbol{\theta}}$ - encoder pretrained via CMTSSL.

\hspace{-2.2em} {\color{commentGreen}\# Compute average gradient magnitude for each image}
\State $\mathcal{G} \gets \emptyset$
\For{$i = 1$ to $N$}
    \State $\mathbf{G}_x \gets \mathbf{I}_i \odot \mathbf{K}_x$, $\mathbf{G}_y \gets \mathbf{I}_i \odot \mathbf{K}_y$, $\mathbf{G}_z \gets \mathbf{I}_i \odot \mathbf{K}_z$
    \State $G_i \gets \texttt{average}\!\left( \!\sqrt{\mathbf{G}_x^{*2} + \mathbf{G}_y^{*2} + \mathbf{G}_z^{*2}}\right)$
    \State $\mathcal{G} \gets \mathcal{G} \cup \{G_i\}$
\EndFor

\hspace{-2.2em} {\color{commentGreen} \# Sort images based on gradient magnitude (ascending)}
\State $ \mathcal{D} \gets \texttt{sort} \left(\mathcal{D}, \texttt{criterion}=\mathcal{G}, \texttt{ord}=\mbox{'asc'}  \right)$

\hspace{-2.2em} {\color{commentGreen} \# Curriculum multi-task self-supervised learning}
\State $E \gets K$

\For{batch $k = 1$ \textbf{to} $S$}

    \hspace{-0.8em} {\color{commentGreen} \# Select data for batch $k$}

    \State $R \gets \left\lfloor N \cdot \tfrac{k}{S} \right\rfloor$
    \State $\mathcal{D}_k \gets \{ \mathbf{I}_i \in  \mathcal{D} \mid i \leq R \}$

    \For{epoch $e = 1$ \textbf{to} $E$}
        \For{$\mathbf{I} \in \mathcal{D}_k$}
        
            \hspace{2.2em} {\color{commentGreen} \# Set up each pretext task}
            \State $(\mathbf{I}^{\text{spa}}, \mathbf{y}^{\text{spa}}) \gets \texttt{spatialJigsaw}(\mathbf{I})$
            \State $(\mathbf{I}^{\text{spe}}, \mathbf{y}^{\text{spe}}) \gets \texttt{spectralJigsaw}(\mathbf{I})$
            \State $(\mathbf{I}^{\text{vis}}, \mathbf{I}^{\text{mask}}) \gets \texttt{masking}(\mathbf{I}, M)$

            \hspace{2.2em} {\color{commentGreen} \# Compute predictions}
            \State $\mathbf{\hat{y}}^{\text{spa}} \gets  h_{\boldsymbol{\phi}}^{\text{spa}}(f_{\boldsymbol{\theta}}(\mathbf{I}^{\text{spa}}))$
            \State $\mathbf{\hat{y}}^{\text{spe}} \gets h_{\boldsymbol{\beta}}^{\text{spe}}(f_{\boldsymbol{\theta}}(\mathbf{I}^{\text{spe}}))$
            \State $\mathbf{\hat{I}}^{\text{mask}} \gets h_{\boldsymbol{\gamma}}^{\text{mim}}(f_{\boldsymbol{\theta}}(\mathbf{I}^{\text{vis}}))$
            
            \hspace{2.2em} {\color{commentGreen} \# Compute losses}
            \State $\mathcal{L}_{\text{spa}} = \text{BCE}\left(\mathbf{y}^{\text{spa}}, \mathbf{\hat{y}}^{\text{spa}}\right)$
            
            \State $\mathcal{L}_{\text{spe}} = \text{BCE}\left(\mathbf{y}^{\text{spe}}, \mathbf{\hat{y}}^{\text{spe}}\right)$
            
            \State $\mathcal{L}_{\text{mim}} = \text{MAE}\left(\mathbf{{I}}^{\text{mask}}, \mathbf{\hat{I}}^{\text{mask}} \right)$

            \State $\mathcal{L}_{\text{total}} = \alpha_{\text{spa}} \cdot\mathcal{L}_{\text{spa}} + \alpha_{\text{spe}} \cdot\mathcal{L}_{\text{spe}} + \alpha_{\text{mim}} \cdot\mathcal{L}_{\text{mim}}$
            
            \hspace{2.2em} {\color{commentGreen} \# Update weights via backpropagation}
            \State $\boldsymbol{\phi} \gets \boldsymbol{\phi} - \eta \cdot \nabla_{\!\boldsymbol{\phi}}\, \mathcal{L}_{\text{spa}}$
            \State $\boldsymbol{\beta} \gets \boldsymbol{\beta} - \eta \cdot \nabla_{\!\boldsymbol{\beta}}\, \mathcal{L}_{\text{spe}}$
            \State $\boldsymbol{\gamma} \gets \boldsymbol{\gamma} - \eta \cdot \nabla_{\!\boldsymbol{\gamma}}\, \mathcal{L}_{\text{mim}}$
            \State $\boldsymbol{\theta} \gets \boldsymbol{\theta} - \eta \cdot \nabla_{\!\boldsymbol{\theta}}\, \mathcal{L}_{\text{total}}$ 
        \EndFor
    \EndFor
    \State $E \gets E \cdot F$
\EndFor
\end{algorithmic}
}
\end{algorithm}  

\vspace{-0.1cm}
\subsection{Curriculum Learning via Gradient Magnitudes}
\label{sec:cMTSS}

\subsubsection{Intuition and empirical motivation} 
To quantify the visual difficulty (complexity) of input images, we introduce a criterion based on image gradients. Our approach leverages the intuition that images with higher gradient magnitudes are more salient, containing richer spatial-spectral structures, e.g.~sharp edges, textures, or abrupt spectral transitions, and are therefore more challenging to model than smooth (homogeneous) images.
To validate our intuition, we conduct preliminary experiments on the large-scale HYPSO dataset \cite{justo_open_2023} and measure the correlation between image gradients and loss values for the masked image modeling and jigsaw puzzle solving tasks. At the same time, we consider various ways to aggregate image gradients and obtain a representative value at the image level, namely via computing the average, standard deviation, or maximum values. 
The correlations reported in Table~\ref{tab:correlation_curri} confirm our intuition, namely that gradient magnitudes are well correlated to SSL task objectives, regardless of the aggregation operator. The results also indicate that the \texttt{average} operator is more correlated with both task losses than the other operators. We thus employ the average gradient magnitude in our framework.

\subsubsection{Gradient computation} 
For each image $\mathbf{I}$, we compute local gradients along both spatial and spectral dimensions. Spatial gradients are estimated using Scharr operators \cite{jahne1999principles} in the horizontal ($x$) and vertical ($y$) directions, while spectral differences between adjacent channels provide a proxy for inter-band variability:
\begin{equation}
    \mathbf{G}_x = \mathbf{I} \odot \mathbf{K}_x, \quad 
    \mathbf{G}_y = \mathbf{I} \odot \mathbf{K}_y, \quad 
    \mathbf{G}_z = \mathbf{I} \odot \mathbf{K}_z, \quad 
\end{equation}
where $\mathbf{K}_x$ and $\mathbf{K}_y$ are Scharr kernels, and $\mathbf{K}_z$ denotes a kernel that performs the difference on the spectral axis ($z$), i.e.~$[1, -1]$. The average gradient magnitude for image $\mathbf{I}$ is given by:
\begin{equation}
    G = \texttt{average}\!\left( \!\sqrt{\mathbf{G}_x^{*2} + \mathbf{G}_y^{*2} + \mathbf{G}_z^{*2}}\right).
\end{equation}
The average gradients magnitudes are computed in steps 1-6 of Algorithm \ref{alg:cmtssl}.

\setlength{\textfloatsep}{\textfloatsepsave}
\begin{table}[t!]
\centering
\caption{Pearson correlation coefficients between individual task losses (MIM and JPS) and difficulty criteria computed on the HYPSO \cite{justo_open_2023} validation set. Gradient magnitudes are positively correlated with SSL task losses, regardless of the gradient aggregation operator.}
\vspace{-0.1cm}
\label{tab:correlation_curri}
\begin{tabular}{|l|c|c|}
\hline
Gradient aggregation  & MIM  & JPS \\
\hline
Average  & $\bm{0.847}$   & $\bm{0.340}$\\
Maximum & $0.742$ &$0.241$ \\
Standard Deviation & $0.521$  & $0.227$ \\
\hline
\end{tabular}
\vspace{-1.5em}
\end{table}

\subsubsection{Data organization and curriculum schedule}  

The average gradient magnitude is used as a proxy for image difficulty in our curriculum learning framework. Images with low average gradient magnitude values (smooth, homogeneous content) are considered \emph{easy}, while those with high gradient magnitude (salient content, complex patterns) are considered \emph{hard}. We sort the samples in ascending order according to their average gradient magnitudes (step 7 of Algorithm \ref{alg:cmtssl}). This ordering allows the model to first capture global spectral-spatial regularities, before tackling more challenging images containing fine-grained or high-frequency structures.

We next divide the sorted samples into $S$ curriculum batches\footnote{Not to be confused with mini-batches. A curriculum batch is much larger, containing multiple training mini-batches.}. Let $\mathcal{D}$ denote the sorted dataset of $N$ training images computed in step 7. Each batch $\mathcal{D}_k$ contains all images $\mathbf{I}_i\!\in\! \mathcal{D}$, such that $i\!\leq\! R$ (step 11), where $R\!=\!\left\lfloor N \cdot \tfrac{k}{S} \right\rfloor$ (step 10). This selection procedure ensures that, in early stages, the training batches contain the easiest images, while harder images are progressively introduced until the entire dataset is covered. Notice that $\mathcal{D}_k\subset \mathcal{D}_{k+1}, \forall k\! <\! S$, which is required to avoid catastrophic forgetting \cite{kirkpatrick2017overcoming}.

We partition the total training budget (expressed in epochs) across the $S$ curriculum batches, according to a schedule parameterized by the number of epochs $K \in \mathbb{N}^+$ for the first batch (step 8) and the growth factor $F\!>\!0$ (step 30). Thus, the number of epochs on batch $\mathcal{D}_k$ is given by $E=K\cdot F^{k-1}$, where $k$ is the curriculum batch index.

\vspace{-0.1cm}
\subsection{Multi-Task Learning}
\vspace{-0.1cm}

We propose an SSL framework that comprises three tasks: spatial jigsaw puzzle solving, spectral jigsaw puzzle solving, and MIM. Jigsaw puzzle solving (JPS) \cite{noroozi-eccv-2016} is a self-supervised task that produces semantically relevant representations. Inspired by Wang et al.~\cite{Wang_video_2022}, who decoupled the JPS task across spatial and temporal dimensions to detect abnormal events in video, we adapt the decoupled JPS version to capture both spatial and spectral structures in hyperspectral imagery. While JPS encourages features that capture contextual relationships between image parts, MIM promotes features that capture fine-grained local details. Since these SSL tasks focus on different aspects, it makes sense to try to combine them to obtain more robust representations. We next present how each task is constructed and how the whole architecture is integrated.

\subsubsection{Spatial jigsaw puzzle solving}
\label{sec:jigsaw}

Each input image is split into non-overlapping 3D patches across the spatial dimensions, such that the size of a 3D patch is ${H_{\text{spa}}\! \times\! W_{\text{spa}}\! \times\! C}$, where $H_{\text{spa}}\!<\!H$ and $W_{\text{spa}}\!<\!W$, respectively. The patches are permuted and reassembled into an output image denoted as $\mathbf{I}^{\text{spa}}$, while the permutation indices are used to compute the target labels for the pretext classification task. Let $N_{\text{spa}} = \frac{H}{H_{\text{spa}}} \cdot \frac{W}{W_{\text{spa}}}$ denote the number of 3D patches. To avoid the combinatorial explosion associated with classifying permutations into different classes, we frame the spatial JPS task as a multi-label classification problem, i.e.~the position of each 3D patch needs to be independently predicted. This essentially reduces the number of output probabilities from $N_{\text{spa}}!$ to $N_{\text{spa}}^2$. The multi-label targets are stored in $\mathbf{y}^{\text{spa}}$. The procedure to generate the spatial JPS task is executed in step 14 of Algorithm \ref{alg:cmtssl}.

\subsubsection{Spectral jigsaw puzzle solving}
\label{sec:spec_jigsaw}

For the spectral JPS task, we permute contiguous spectral blocks, where the size of each block is ${H \times W \times C_{\text{spe}}}$, with $C_{\text{spe}}\!<\!C$. The number of spectral blocks $N_{\text{spe}}$ is given by $N_{\text{spe}} = \frac{C}{C_{\text{spe}}}$. We adopt the same multi-label classification formulation as before, reducing the factorial number of classes to $N_{\text{spe}}^2$. The procedure to generate the spectral JPS task executed in step 15 returns the shuffled input $\mathbf{I}^{\text{spe}}$ and the target labels $\mathbf{y}^{\text{spe}}$.

\subsubsection{Masked image modeling}
\label{sec:masking}
For the MIM task, each hyperspectral image $\mathbf{I}$ is divided into non-overlapping 3D patches, such that the size of a patch is ${H_{\text{mim}} \!\times\! W_{\text{mim}} \!\times\! C_{\text{mim}}}$, where $H_{\text{mim}}\!<\!H$,  $W_{\text{mim}}\!<\!W$ and $C_{\text{mim}}\!<\!C$, respectively. A randomly chosen subset of $M$ patches is masked, resulting in a visible version of the input, denoted as $\mathbf{I}^{\text{vis}}$, and a masked version, denoted as $\mathbf{I}^{\text{mask}}$ (step 16). The visible patches $\mathbf{I}^{\text{vis}}$ are given as input to the model, which is supposed to reconstruct the masked patches $\mathbf{I}^{\text{mask}}$.

\subsubsection{Multi-task learning architecture}
\label{sec:MTSS}

Our architecture comprises a shared encoder and three task-specific heads (see Figure~\ref{fig:full_pipeline}). The shared encoder $f_{\boldsymbol{\theta}}$ learns a joint representation space for all SSL tasks. One head, denoted as $h_{\boldsymbol{\gamma}}^{\text{mim}}$, is dedicated to the MIM task, while the other two, $h_{\boldsymbol{\phi}}^{\text{spa}}$ and $h_{\boldsymbol{\beta}}^{\text{spe}}$, correspond to the spatial and spectral JPS tasks. 

Each task generation procedure takes the training image $\mathbf{I}$ as input and returns the task-specific input and target (steps 14-16). The resulting inputs are then passed through the shared encoder and routed to their respective task-specific heads. Each head produces a task-specific output (steps 17-19), which is compared with the corresponding target via a task-specific loss function (steps 20-22). The spatial and spectral JPS losses, denoted as $\mathcal{L}_{\text{spa}}$ and $\mathcal{L}_{\text{spe}}$, predict the corresponding permutations via binary cross-entropy (BCE). The reconstruction loss $\mathcal{L}_{\text{mim}}$ is computed as the mean absolute error (MAE) between predicted and ground-truth pixel values in the masked regions. The three losses are subsequently combined via a weighted sum (step 23), modulated by the hyperparameters $\alpha_{\text{spa}}, \alpha_{\text{spe}}, \alpha_{\text{mim}}$, respectively.  
The final training objective is defined as follows:
\vspace{-1pt}
\begin{equation}
  \mathcal{L}_{\text{total}} = \alpha_{\text{spa}} \cdot \mathcal{L}_{\text{spa}} +\alpha_{\text{spe}} \cdot \mathcal{L}_{\text{spe}} + \alpha_{\text{mim}} \cdot \mathcal{L}_{\text{mim}}.
\end{equation}

We update the parameters $\boldsymbol{\phi}$, $\boldsymbol{\beta}$, $\boldsymbol{\gamma}$ of the task-specific heads via the corresponding loss functions (steps 24-26). The shared encoder $f_{\boldsymbol{\theta}}$ is trained by back-propagating gradients from all three heads via $\mathcal{L}_{\text{total}}$ (step 27).

Upon completing the pretraining phase via CMTSSL, the pretrained encoder $f_{\boldsymbol{\theta}}$ can be fine-tuned on downstream tasks by attaching an appropriate task-specific head. 

\section{Experiments}

We conduct segmentation (a.k.a.~pixel-wise classification) experiments to compare single-task and multi-task SSL, and to assess the benefits of curriculum learning in the context of multi-task SSL. We also compare the lightweight models trained via CMTSSL with state-of-the-art models, including both lightweight architectures and computationally-intensive foundation models. Finally, we perform ablation studies on various hyperparameters to complement the main results.

\vspace{-0.1cm}
\subsection{Datasets}
\vspace{-0.1cm}

The \emph{HYPSO-1 ``Sea-Land-Cloud''} dataset \cite{justo_open_2023} comprises 179 hyperspectral images captured by the HYPSO-1 satellite across diverse global locations. Among these, 38 images include pixel-level annotations, providing approximately 25 million labeled spectral signatures for the sea, land, and cloud classes. The remaining 141 unlabeled images support unsupervised and self-supervised learning. Each image has a spatial resolution of $956 \times 684$ pixels with 120 spectral bands (400-800\texttt{nm}). The official split consists of 30 training, 3 validation, and 5 test images. 

The \emph{Pavia University (PU)} and \emph{Pavia Center (PC)} datasets contain hyperspectral images acquired by the ROSIS sensor during an aerial survey over Pavia, northern Italy. Both provide a spatial resolution of 1.3\texttt{m} per pixel and include ground-truth annotations for nine land-cover types and a background class. PU consists of a single scene of $610 \times 340$ pixels with 42,776 labeled pixels ($20\%$ of total pixels) across 103 spectral bands (430-860\texttt{nm}). PC comprises a scene of $1096 \times 715$ pixels with 148,152 labeled pixels ($19\%$) across 102 bands in the same range.

The \emph{WHU-Hi-HanChuan (HC)} dataset is part of the WHU-Hi benchmark suite released by the RSIDEA group at Wuhan University. 
It was captured using an UAV-mounted hyperspectral camera at 250\texttt{m} altitude. The scene contains $1217 \times 303$ pixels at 0.109\texttt{m} spatial resolution, with 274 spectral bands spanning 400-1000\texttt{nm}. It includes annotations for 16 distinct classes. 

\begin{table}[t!]
\centering
\caption{Number of subimages per split for all datasets.}
\vspace{-5pt}
\label{tab:dataset_info}
\begin{tabular}{|l|c|c|c|c|}
\hline
Dataset  & Pretraining  & Training & Validation & Testing \\
\hline
PU & 1,653 & 180 & 90 & 24,873 \\
PC & 6,364 & 180 & 90 & 90,146   \\
HC & 1,681 & 800 & 400 &171,652 \\
HYPSO & 441,180 & 77,400 & 7,740 & 3,269,520  \\
\hline
\end{tabular}
\vspace{-1.5em}
\end{table}
\vspace{-0.1cm}
\subsection{Experimental Setup}
\vspace{-0.1cm}

\subsubsection{Data preprocessing}
\label{sec:Data preprocessing}

Following standard preprocessing in previous work \cite{justo_semantic_2025}, we divide the original high-resolution images from all datasets into subimages of $16\times16$ pixels, while preserving the number of channels (spectral bands). To ensure consistent input distributions and stabilize training, we normalize the data by computing the mean and standard deviation for each spectral band. For the three single-scene datasets (PU, PC and HC), we partition each scene into non-overlapping regions. These regions are distributed into training, validation, and test sets, to prevent data leakage during evaluation. For pretraining, we sample additional $16\times16$ subimages from the training regions, using an overlap (stride) of $8$ pixels. In Table~\ref{tab:dataset_info}, we report the resulting number of subimages (later called images) for each dataset.

\subsubsection{Chosen architectures}
We evaluate our CMTSSL framework on three lightweight architectures: 2D Justo-UNet-Simple (2D Justo)~\cite{justo_semantic_2025}, 2D CUNet++ Reduced \cite{justo_semantic_2025}, and CLOLN \cite{li2024channel}. The first two architectures are the best and second-best 2D CNN models on the HYPSO dataset~\cite{justo_open_2023}. All chosen architectures are suitable for onboard processing, containing between 4K and 11K learnable parameters.

\subsubsection{State-of-the-art models} Aside from the chosen architectures, we compare with additional models, namely the HyperSIGMA-B foundation model \cite{wang_hypersigma_2025}, as well as the lightweight 1D Justo-LiuNet \cite{justo_semantic_2025} and FastViT-S12 \cite{anasosalu_vasu_fastvit_2023}.

\begin{table}[t!]
\caption{State-of-the-art comparison on the Pavia University, Pavia Center, WHI-Hancuan and HYPSO datasets. CMTSSL is applied on 2D Justo, CUNet++ Reduced, and CLOLN, respectively. The best score on each dataset is highlighted in {\color{Green}\textbf{bold green}}. $*$ -- results are taken from \cite{justo_semantic_2025}.}
\label{tab:global}
\vspace{-5pt}
\setlength\tabcolsep{0.35em}
\centering
  \begin{tabular}{|c|l|c|c|c|}
  \hline
    & Model & AA & Parameters & FLOPs \\
    \hline
    \multirow{8}{*}{\rotatebox{90}{Pavia University}} & HyperSIGMA-B \cite{wang_hypersigma_2025} & $62.2\pm2.0$ &  $176,700K$& $66,800,336 K$\\
    & 1D Justo-LiuNet \cite{justo_semantic_2025} &$65.6	\pm2.1$ & $4.6K$ &$51K$\\
    \cline{2-5}
    & 2D Justo \cite{justo_semantic_2025} &$73.5	\pm3.3$ & $9K$ & $1,494K$\\
    & \quad+CMTSSL (ours) &$75.8	\pm0.3$ & $9K$ & $1,494K$\\
    & CUNet++ Reduced \cite{justo_semantic_2025}&$59.8\pm1.8$ & $11K$ & $1,949K$\\
    & \quad+CMTSSL (ours) &$68.1\pm1.5$ & $11K$ & $1,949K$\\
    & CLOLN \cite{li2024channel} & $78.2\pm 5.6$ & $3.5K$ & $926K$\\
    & \quad+CMTSSL (ours) & $\best{79.4\pm 2.5}$ & $3.5K$ & $926K$\\
    \hline
    \multirow{8}{*}{\rotatebox{90}{Pavia Center}}     & HyperSIGMA-B \cite{wang_hypersigma_2025} & $69.3\pm5.5$ &  $176,700K$& $66,800,336 K$\\
    &
    1D Justo-LiuNet \cite{justo_semantic_2025} &$76.8	\pm0.3$ & $4.6K$ &$51K$\\
    \cline{2-5}
    & 2D Justo \cite{justo_semantic_2025} &$79.1	\pm6.8$ & $9K$ & $1,494K$\\
    & \quad+CMTSSL (ours) &$83.0	\pm8.0$ & $9K$ & $1,494K$\\
    & CUNet++ Reduced \cite{justo_semantic_2025}&$74.1\pm4.7$ & $11K$ & $1,949K$\\
    & \quad+CMTSSL (ours) &$77.3\pm4.2$ & $11K$ & $1,949K$\\
    & CLOLN \cite{li2024channel} & $82.7\pm 2.6$ & $3.5K$ & $926K$\\
    & \quad+CMTSSL (ours) & $\best{85.7\pm 4.2}$ & $3.5K$ & $926K$\\
    \hline
    \multirow{8}{*}{\rotatebox{90}{WHU-HI Hanchuan}} & HyperSIGMA-B \cite{wang_hypersigma_2025} & $\best{64.3\pm2.5}$ &  $176,700K$& $66,800,336 K$\\
    &
    1D Justo-LiuNet \cite{justo_semantic_2025} &$55.0	\pm1.0$ & $9K$ &$206K$\\
    \cline{2-5}
    & 2D Justo \cite{justo_semantic_2025}&$61.1	\pm1.9$ & $18K$ & $3,827K$\\
    & \quad+CMTSSL (ours) &$61.7	\pm1.5$ & $18K$ & $3,827K$\\
    & CUNet++ Reduced \cite{justo_semantic_2025}&$53.5\pm1.3$ & $23K$ & $5,045K$\\
    & \quad+CMTSSL (ours) &$53.6\pm0.2$ & $23K$ & $5,046K$\\
    & CLOLN \cite{li2024channel} & $60.9\pm 0.4$ & $6K$ & $1,614K$\\
    & \quad+CMTSSL (ours) & $63.5\pm 2.4$ & $6K$ & $1,614K$\\
    
    \hline
    \multirow{8}{*}{\rotatebox{90}{HYPSO}} & FastViT-S12$^*$ \cite{anasosalu_vasu_fastvit_2023} & $90.0$ &  $10,243K$& - \\
    &
    1D Justo-LiuNet$^*$ \cite{justo_semantic_2025} &$93.0$ & $5K$ &$59K$\\
       \cline{2-5}
    & 2D Justo \cite{justo_semantic_2025}&$92.9 \pm 0.3$ & $9K$ & $1,642K$\\
    & \quad+CMTSSL (ours) &$\best{93.5\pm0.2}$ & $9K$ & $1,642K$\\
    & CUNet++ Reduced \cite{justo_semantic_2025}&$92.8\pm0.6$ & $12K$ & $2,132K$\\
    & \quad+CMTSSL (ours) &$92.9\pm0.8$ & $12K$ & $2,132K$\\
        & CLOLN \cite{li2024channel} & $91.9\pm 0.1$ & $4K$ & $967K$\\
    & \quad+CMTSSL (ours) & $92.4\pm 0.3$ &  $4K$ & $967K$\\
    \hline
    \end{tabular}
\vspace{-0.4cm}
\end{table}

\subsubsection{Hyperparameter tuning}
All models are optimized with AdamW using a learning rate of $5\cdot10^{-4}$ and a mini-batch size of $16$. Each model is pretrained for 120-250 epochs, depending on the dataset. The masking ratio ($0.6$) and patch dimensions are kept consistent across all experiments. The loss weights in the CMTSSL framework are carefully tuned to balance the optimization of individual objectives. More specifically, they are scaled to ensure that the gradient magnitudes of each loss component remain comparable, thereby preventing any single objective from disproportionately influencing the training process. Curriculum hyperparameters ($S$, $K$ and $F$) are set in a dataset-dependent manner, considering the following ranges: $S \in \{3, 4, 5\}$, $K \in \{10,11,...,40\}$, $F \in \{1, 1.1, ..., 2 \}$. They are chosen such that the total number of batches processed during pretraining remains consistent between the curriculum-based and non-curriculum setups, ensuring a fair comparison. We tune hyperparameters using grid-search on validation data.

\begin{table*}[t]
\caption{Results of lightweight 2D Justo, CUNet++ and CLOLN encoders trained from scratch or pretrained via MIM, JPS, MTSSL, and CMTSSL, on the Pavia University, Pavia Center, WHI-Hancuan and HYPSO datasets. The best and second-best scores of each model on each dataset are highlighted in {\color{Green}\textbf{bold green}} and {\color{blue}\underline{underline blue}}, respectively.}
\label{tab:results}
\vspace{-0.3cm}
\setlength\tabcolsep{0.34em}
\begin{center}
  \begin{tabular}{|cc|l|ccc|ccc|ccc|}
  \hline
   \multicolumn{2}{|c|}{Data} & Training  & \multicolumn{3}{c|}{{2D Justo \cite{justo_semantic_2025}}} & \multicolumn{3}{c|}{{CUNet++ Reduced \cite{justo_semantic_2025}}} & \multicolumn{3}{c|}{{CLOLN \cite{li2024channel}}}\\
    \cline{4-12}
    \multicolumn{2}{|c|}{set} & strategy & AA & OA & Kappa & AA & OA & Kappa & AA & OA & Kappa \\
    \hline
    
    \multirow{5}{*}{\rotatebox{90}{Pavia}} & \multirow{5}{*}{\rotatebox{90}{University}}  & From scratch   & $73.5\pm3.3$&$59.3\pm4.8$&$51.5\pm4.8$&                     $59.8\pm1.8$&$53.0\pm4.4$&$43.2\pm4.7$&                           $\second{78.2\pm4.3}$&$\second{70.3\pm5.6}$&$\second{64.1\pm7.0}$\\
    & & MIM       & $\second{75.5\pm1.8}$	&$\second{64.4\pm1.6}$&$\second{56.7\pm2.0}$&                  $62.4\pm1.0$&$52.0\pm1.7$&$42.5\pm1.7$&                           $74.7\pm2.6$&$63.5\pm2.7$&$55.8\pm3.4$\\
    & & JPS       & $73.8\pm	1.1$&$61.3\pm3.0$&$53.9\pm3.0$&                $66.5\pm3.3$	&$\second{57.0\pm3.0}$&                            $\second{48.3\pm3.3}$&$77.0\pm4.0$&$67.8\pm6.0$&$61.4\pm7.0$\\
    & & MTSSL   & $72.7\pm	1.9$&$59.7\pm2.8$&$51.6\pm3.2$&                 $\best{68.5\pm1.2}$&$55.4\pm3.6$&$46.7\pm3.4$&                    $75.3\pm3.1$&$68.8\pm2.7$&$61.7\pm3.4$\\
    & & CMTSSL (ours) & $\best{75.8\pm0.3}$&$\best{64.8\pm0.9}$&$\best{57.5\pm0.8}$&       $\second{68.1\pm1.5}$&$\best{57.2\pm1.3}$&$\best{48.8\pm1.4}$&    $\best{79.4\pm2.5}$&$\best{70.6\pm5.8}$&$\best{64.9\pm7.0}$\\    
    \hline

    \multirow{5}{*}{\rotatebox{90}{Pavia}} & \multirow{5}{*}{\rotatebox{90}{Center}}  & From scratch    &$79.1\pm6.8$&$91.6\pm3.8$&$87.5\pm	5.8$&                     $74.1\pm4.7$&$86.1\pm2.3$&$79.4\pm3.4$&                          $82.7\pm2.6$&$93.0\pm2.6$&$89.5\pm4.0$\\
    & & MIM        &$\best{83.7\pm3.4}$&$\best{93.4\pm2.2}$&$\best{90.1\pm3.2}$&                       $\best{77.4\pm3.4}$&$87.8\pm2.1$&$82.0\pm3.0$&                   $83.4\pm0.7$&$\second{93.5\pm1.0}$&$\second{90.3\pm1.4}$\\
    & & JPS        &$81.8\pm2.4$&$92.2\pm1.0$&$88.4\pm1.4$&                       $75.0\pm4.4$&$87.5\pm0.9$&$81.5\pm1.4$&                          $\second{84.0\pm0.2}$&$\second{93.5\pm0.7}$&$\second{90.3\pm0.9}$\\
    & & MTSSL    &$82.9\pm1.9$&$92.1\pm0.7$&$88.3\pm1.0$&             $76.5\pm2.2$&$\second{87.9\pm0.8}$&$\second{82.1\pm1.1}$&        $82.2\pm1.3$ &$92.9\pm1.4$&$89.4\pm2.2$\\
    & & CMTSSL (ours)   &$\second{83.0\pm8.0}$ & $\second{92.4\pm3.9}$&$\second{88.6\pm5.7}$&                  $\second{77.3\pm4.2}$&$\best{88.0\pm2.4}$&$\best{82.2\pm3.6}$&   $\best{85.7\pm4.2}$ & $\best{94.7\pm1.5}$& $\best{92.1\pm2.2}$ \\
    \hline

    \multirow{5}{*}{\rotatebox{90}{WHU-HI}} & \multirow{5}{*}{\rotatebox{90}{Hanchuan}} &
    From scratch  & $61.1\pm1.9$&$73.2\pm1.6$&$69.5\pm1.8$&                                        $\second{53.5\pm1.3}$&$65.0\pm2.0$&$\second{60.3\pm2.1}$&                     $60.9\pm0.4$&$71.9\pm3.7$&$68.0\pm1.3$\\
    & & MIM      & $58.6\pm0.7$ & $72.5\pm1.4$& $68.6\pm1.5$&                                     $52.1\pm1.6$ & $64.3\pm1.6$& $59.6\pm1.7$ &                    ${61.5\pm1.4}$&$71.5\pm0.5$& $67.7\pm0.5$ \\
    & & JPS      & $59.2\pm1.8$ & $69.8\pm3.0$& $65.8\pm3.3$&                                     $51.9\pm2.4$ & $60.5\pm5.4$& $55.3\pm5.7$ &      $61.1\pm5.0$&$70.9\pm4.0$& $66.9\pm4.4$\\
    & & MTSSL  & $\best{62.1\pm1.7}$ & $\best{74.5\pm0.6}$& $\best{70.9\pm0.6}$&                $52.5\pm1.6$ & $\best{66.1\pm0.6}$& $\best{61.7\pm0.7}$ &                      $\second{62.6\pm0.8}$ & $\second{71.9\pm1.4}$& $\second{68.1\pm1.6}$\\
    & & CMTSSL (ours) & $\second{61.7\pm1.5}$ & $\second{73.8\pm1.1}$& $\second{70.1\pm1.1}$&          $\best{53.6\pm0.2}$ & $\second{65.1\pm0.9}$& $60.1\pm1.1$ &                      $\best{63.5\pm2.4}$ & $\best{72.3\pm1.9}$& $\best{68.4\pm2.2}$\\
    \hline
    \multicolumn{2}{|c|}{\multirow{5}{*}{\rotatebox{90}{HYPSO}}} &
    From scratch     &$92.9\pm0.3$&$\second{95.4\pm0.2}$&$92.4\pm0.4$&                                     $92.8\pm0.6$&$94.7\pm0.1$&$91.2\pm0.2$&                          $91.9\pm0.1$&$94.7\pm0.2$&$91.2\pm0.3$\\
    & & MIM         &$92.6\pm0.6$ & $94.8\pm0.0$& $91.4\pm0.1$ &                                 $92.7\pm0.4$&$94.4\pm0.5$&$90.7\pm0.9$&                            $\second{92.3\pm0.9}$&$94.7\pm0.6$&$91.3\pm 0.9$\\
    & & JPS         &$93.0\pm0.2$ & $95.3\pm0.2$& $92.2\pm0.3$ &                                 $\second{92.9\pm0.3}$&$\best{95.3\pm0.3}$&$\best{92.2\pm0.4}$&                       $91.7\pm0.8$ & $\second{95.1\pm0.5}$& $\second{91.8\pm0.8}$\\
    & & MTSSL (ours)     &$\second{93.3\pm0.7}$ & $\best{95.5\pm0.5}$& $\second{92.5\pm0.8}$ &      $\best{93.1\pm0.5}$&$\second{94.9\pm1.0}$&$\second{91.6\pm 1.6}$&                        $91.6\pm1.3$ & $94.8\pm0.4$& $91.4\pm0.8$\\
    & & CMTSSL (ours)    &$\best{93.5\pm0.2}$ & $\best{95.5\pm0.1}$& $\best{92.6\pm0.3}$ &            $\second{92.9\pm0.8}$&$\second{94.9\pm1.0}$&$\second{91.6\pm1.5}$&                      $\best{92.4\pm0.3}$ & $\best{95.2\pm0.4}$& $\best{92.0\pm0.7}$\\
    \hline
    \end{tabular}
\end{center}
\label{table:finetuningresults}
\vspace{-0.5cm}
\end{table*}

\subsubsection{Evaluation metrics}
We evaluate models using three commonly employed metrics in hyperspectral classification: Overall Accuracy (OA), which reflects the proportion of correctly classified pixels; Average Accuracy (AA), the mean per-class accuracy; and the Kappa coefficient, which adjusts OA by accounting for the accuracy expected by chance. These metrics provide a comprehensive assessment of model accuracy. Results are averaged over $3$ runs to ensure statistical reliability.

\subsubsection{Computational environment}
Experiments are carried out on high-performance GPUs (NVIDIA A40 and Quadro RTX 6000) provided by a shared scientific computing facility, enabling consistent evaluation across multiple datasets.

\vspace{-0.1cm}
\subsection{Results}
\vspace{-0.1cm}

\subsubsection{Comparison with state of the art}
In Table~\ref{tab:global}, we compare CMTSSL with various state-of-the-art models in terms of average accuracy, parameter count and FLOPs. For a fair comparison, all models are trained and tested on the same data splits. CMTSSL boosts the performance of all three models that is applied on, namely 2D Justo, CUNet++ Reduced, and CLOLN. Notably, the performance gains come without increasing parameter count or FLOPs. Lightweight models often outperform larger models, indicating that models optimized for onboard processing do not have to compromise on accuracy. On the HYPSO dataset, CMTSSL boosts the average accuracy of 2D Justo to $93.5\%$ ($+0.6\%$), setting a new state-of-the-art score that surpasses the previous best ($93\%$), reported by 1D Justo-LiuNet~\cite{justo_semantic_2025}. Large standard deviations on the three one-scene datasets (PU, PC and HC) are likely due to the small amount of training samples per category (see Table~\ref{tab:dataset_info}).

\subsubsection{Comparison of training procedures}

\begin{figure}[t]
  \centering
  \setlength{\tabcolsep}{3pt}
  \small{
  \begin{tabular}{ccc}
  From scratch & CMTSSL & Ground truth \\
  
    \includegraphics[width=0.145\textwidth]{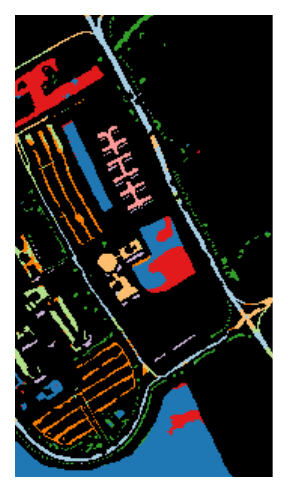} &
    \includegraphics[width=0.145\textwidth]{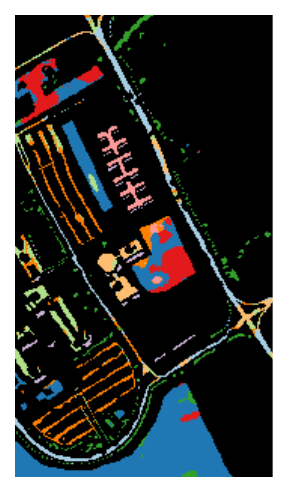} & 
    \includegraphics[width=0.145\textwidth]{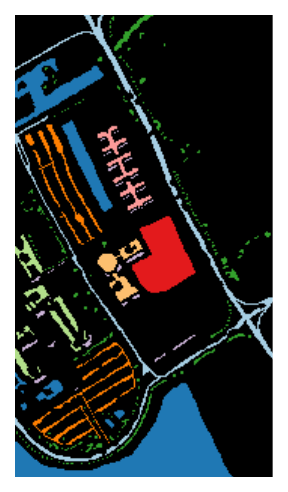} 
  \end{tabular}
  }
  \vspace{-0.28cm}
  \caption{Visual comparison on the Pavia University dataset for the 2D Justo \cite{justo_semantic_2025} architecture, using training from scratch vs.~CMTSSL. Each color represents a different class, with black being the background. Best viewed in color.}
  \label{fig:visual}
\vspace{-0.45cm}
\end{figure}

To assess the benefits of self-supervised pretraining, we compare models trained from scratch (with full supervision) with models that undergo fine-tuning, after pretraining with one the following SSL configurations: MIM only, decoupled JPS only, MTSSL, and CMTSSL (ours). In Table~\ref{tab:results}, we present the comparative evaluation of training procedures. Note that MIM, JPS and MTSSL can be seen as ablated versions of our CMTSSL. The ablated versions are not always able to surpass the training from scratch. In most cases, CMTSSL outperforms the single-task pretraining frameworks across all evaluation metrics. Moreover, CMTSSL is the only self-supervised pretraining framework that is consistently better than training from scratch. The examples shown in Figure \ref{fig:visual} further support our observation. Overall, the results confirm our hypothesis, namely that multi-task self-supervised learning is beneficial, but only with the help of curriculum learning.

\subsubsection{More ablation studies}
\begin{figure*}[!th]
  \centering
  \newlength{\panelheight}
  \setlength{\panelheight}{2.8cm} 
    \setlength{\tabcolsep}{0.5pt}
  \begin{tabular}{ccccc}
    \parbox[c][\panelheight][c]{3mm}{\centering\tiny\rotatebox{90}{Overall accuracy}}&

    \parbox[c][\panelheight][c]{0.23\textwidth}{%
      \includegraphics[width=\linewidth,height=\panelheight]{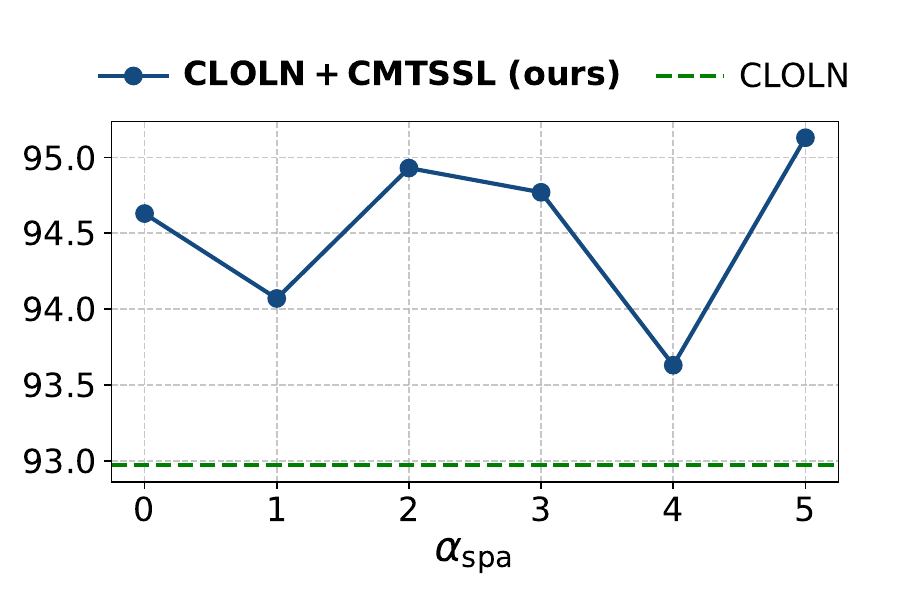}
    } &

    \parbox[c][\panelheight][c]{0.23\textwidth}{%
      \includegraphics[width=\linewidth,height=\panelheight]{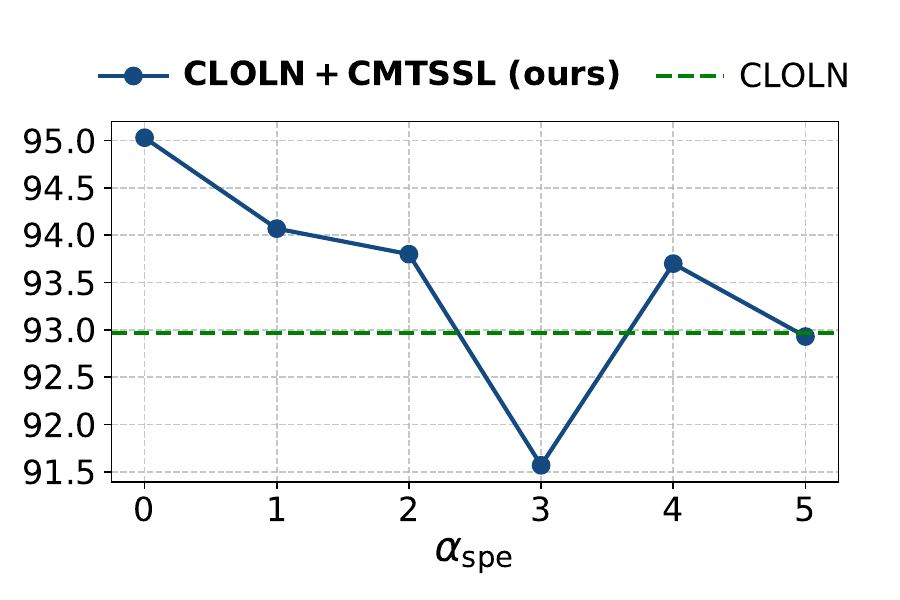}
    } &

    \parbox[c][\panelheight][c]{0.23\textwidth}{%
      \includegraphics[width=\linewidth,height=\panelheight]{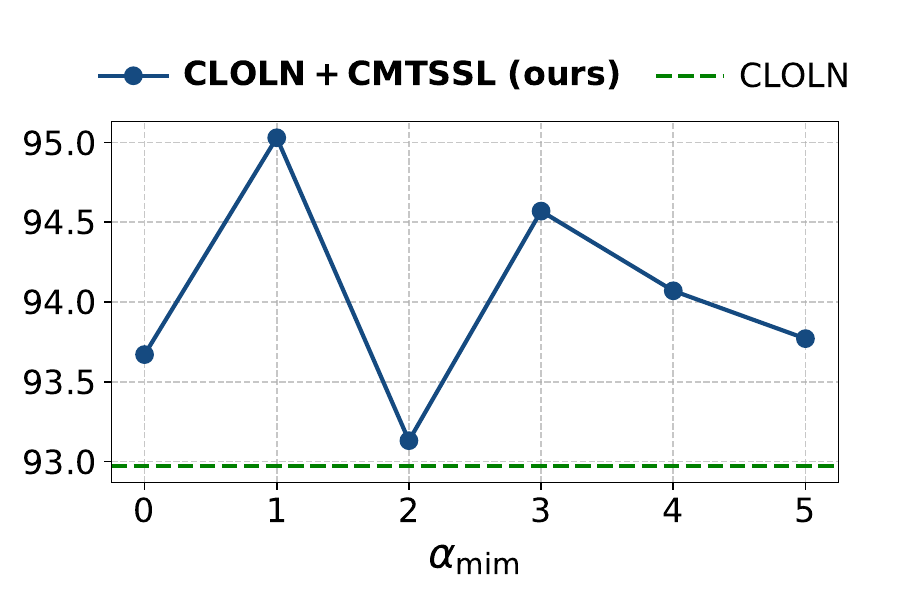}
    } &

    \parbox[c][\panelheight][c]{0.23\textwidth}{%
      \centering\scriptsize
      \setlength\tabcolsep{0.3em}
      \begin{tabular*}{\linewidth}{@{\extracolsep{\fill}}|c|c|}
      \hline
  {Hyperparameters} & {OA} \\
  \hline
  - & $93.0 \pm 2.6$\\
  \hline
  $K\!=\!18$, $F\!=\!1.9$, $S\!=\!3$ & $94.7\!\pm\!2.2$\\
  $K\!=\!32$, $F\!=\!1.5$, $S\!=\!3$ & $94.0\!\pm\!1.0$\\
  $K\!=\!64$, $F\!=\!1.1$, $S\!=\!3$ & $94.3\!\pm\!1.4$\\
  \hline
  $K\!=\!52$, $F\!=\!1.5$, $S\!=\!2$ & $93.3\!\pm\!1.0$\\
  $K\!=\!20$, $F\!=\!1.5$, $S\!=\!4$ & $95.1\!\pm\!1.7$\\
  $K\!=\!13$, $F\!=\!1.5$, $S\!=\!5$ & $94.8\!\pm\!1.0$\\
  \hline
\end{tabular*}
    }
  \end{tabular}
\vspace{-0.2cm}
  \caption{Parameter sensitivity analysis on the Pavia Center dataset using the CLOLN architecture. Base values are $\alpha_{\text{spa}}=1$, $\alpha_{\text{spe}}=1$, $\alpha_{\text{mim}}=4$, $K=32$, $F=1.5$, and $S=3$. Each plot varies a single hyperparameter, while the others remain fixed. The baseline corresponds to CLOLN without CMTSSL. Best viewed in color.}
  \label{fig:ablation_study}
  \vspace{-0.50cm}
\end{figure*}

To assess the influence of individual hyperparameters, we conduct a sensitivity analysis by varying CMTSSL-specific parameters ($\alpha_{\text{spa}}$, $\alpha_{\text{spe}}$, $\alpha_{\text{mim}}$, $K$, $F$, $S$), one at a time. The results shown in Figure~\ref{fig:ablation_study} indicate that CMTSSL is a robust framework, consistently outperforming the baseline model (trained from scratch) across various hyperparameter configurations, demonstrating its robustness and stability.
 
\section{Conclusion}

In this paper, we introduced CMTSSL, a curriculum multi-task self-supervised learning framework tailored for HSI analysis with lightweight architectures. By jointly integrating MIM and decoupled JPS via curriculum learning based on 3D gradient magnitudes, CMTSSL enables efficient and robust representation learning, without increasing model complexity or requiring additional labeled data. Extensive experiments on four benchmark datasets demonstrate that CMTSSL consistently improves downstream segmentation performance across various encoder architectures. Our ablation results confirm the utility of the integrated components. By enabling lightweight models to acquire semantically rich and generalizable features through a unified and efficient pretraining pipeline, CMTSSL provides a promising foundation for next-generation remote sensing systems requiring fast, accurate, and compact hyperspectral processing.

\bibliographystyle{IEEEtran}
\bibliography{biblio}

\end{document}